\pdfoutput=1

\documentclass[11pt]{article}

\usepackage{EMNLP2022}

\usepackage{times}
\usepackage{latexsym}

\usepackage[T1]{fontenc}

\usepackage[utf8]{inputenc}

\usepackage{microtype}

\usepackage{graphicx}
\usepackage{xcolor}

\usepackage{amsmath}
\usepackage{amssymb}

\usepackage{enumerate}

\usepackage{booktabs}
\usepackage{multirow}

\usepackage{makecell}

\usepackage{url}

\newcommand\strudelemoji{\raisebox{-3.8pt}{\includegraphics[width=1.5em]{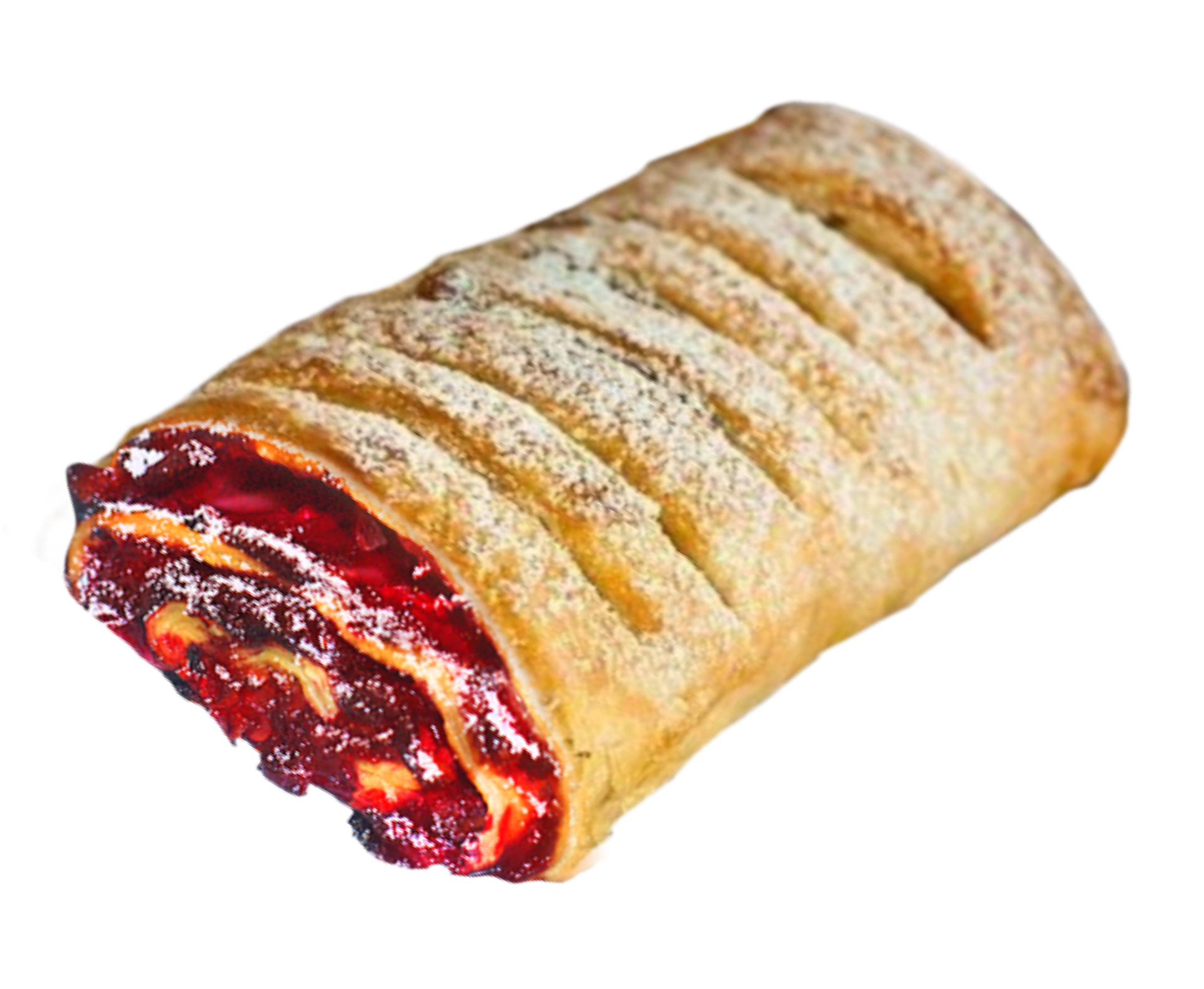}}}

\definecolor{blue1}{RGB}{0, 203, 255}
\definecolor{yellow1}{RGB}{241, 196, 15}

\title{STRUDEL: Structured Dialogue Summarization for\\Dialogue Comprehension}

\author{Borui Wang$^1$ \quad Chengcheng Feng$^1$ \quad Arjun Nair$^1$ \quad Madelyn Mao$^1$ \quad \textbf{Jai Desai}$^1$\\\textbf{Asli Celikyilmaz}$^2$ \quad \textbf{Haoran Li}$^2$ \quad \textbf{Yashar Mehdad}$^2$ \quad \textbf{Dragomir Radev}$^1$\\
$^1$Yale University \quad $^2$Meta AI\\
\scalebox{0.86}[0.9]{\tt{\{borui.wang, dragomir.radev\}@yale.edu} \texttt{\{aslic, aimeeli, mehdad\}@fb.com}}
}

\begin{document}
\maketitle
\begin{abstract}

Abstractive dialogue summarization has long been viewed as an important standalone task in natural language processing, but no previous work has explored the possibility of whether abstractive dialogue summarization can also be used as a means to boost an NLP system's performance on other important dialogue comprehension tasks. In this paper, we propose a novel type of dialogue summarization task - STRUctured DiaLoguE Summarization (STRUDEL\strudelemoji{}) - that can help pre-trained language models to better understand dialogues and improve their performance on important dialogue comprehension tasks. We further collect human annotations of STRUDEL summaries over 400 dialogues and introduce a new STRUDEL dialogue comprehension modeling framework that integrates STRUDEL into a graph-neural-network-based dialogue reasoning module over transformer encoder language models to improve their dialogue comprehension abilities. In our empirical experiments on two important downstream dialogue comprehension tasks - dialogue question answering and dialogue response prediction - we show that our STRUDEL dialogue comprehension model can significantly improve the dialogue comprehension performance of transformer encoder language models.

\end{abstract}

\section{Introduction}

In natural language processing, abstractive dialogue summarization \citep{chen-2021-dialogsum, gliwa-etal-2019-samsum,fabbri-etal-2021-convosumm,zhu-etal-2021-mediasum,zhong-etal-2021-qmsum} has long been viewed as an important standalone task, but no previous work has explored the possibility of whether abstractive dialogue summarization can also be used as a means to boost an NLP system's performance on other important dialogue comprehension tasks. When performing language understanding, a very natural and effective first step that human beings usually take in their mental process is to try to summarize the main content of a piece of text, usually from multiple perspectives each focusing on a different aspect of the text. This is especially true when human readers or speakers are trying to understand a dialogue or a conversation, which involves a multi-turn exchange of information following a general theme, topic, or storyline. Therefore, we would like to ask the following question - can the task of abstractive dialogue summarization also help NLP models to learn to better perform dialogue comprehension tasks?

\begin{figure}[t]
  \centering
  \includegraphics[width = \columnwidth]{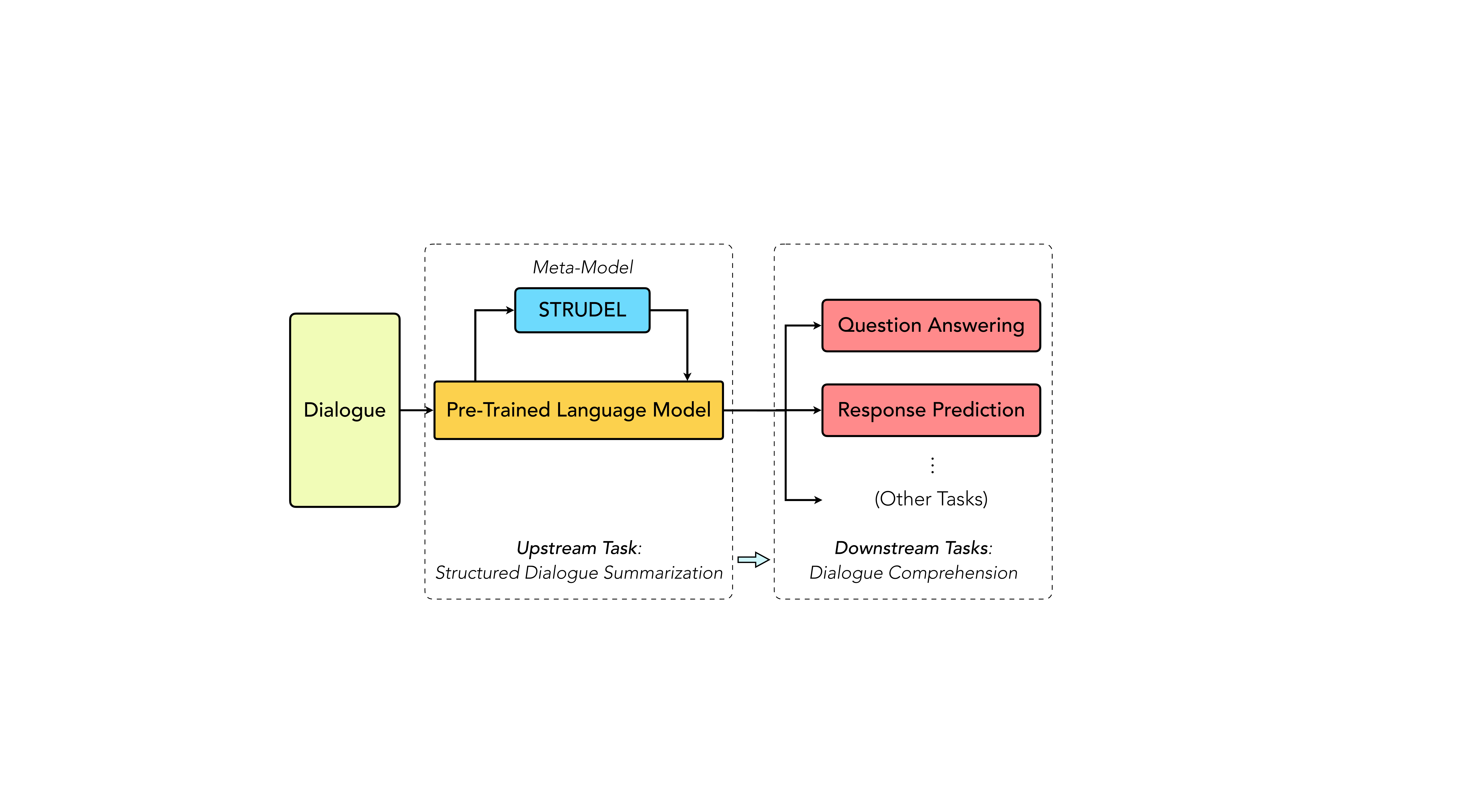}
  \caption{STRUDEL as a meta-model on top of pre-trained language models for dialogue comprehension.}
  \label{fig:meta_model}
\end{figure}

In this paper, we propose a novel type of dialogue summarization task - STRUctured DiaLoguE Summarization (STRUDEL\strudelemoji{}\footnote{The name STRUDEL comes from a type of layered pastry with fillings called \textit{strudel}, which, like our proposed task of structured dialogue summarization, is also structured. }) - that can help pre-trained language models to better understand dialogues and improve their performance on important dialogue comprehension tasks. In contrast to the holistic approach taken by the traditional free-form abstractive summarization task for dialogues, STRUDEL aims to decompose and imitate the hierarchical, systematic, and structured mental process that we human beings usually go through when understanding and analyzing dialogues, and thus has the advantage of being more focused, specific and instructive for dialogue comprehension models to learn from. And guided by our proposed STRUDEL task, we also collect a set of human annotations of STRUDEL summaries over 400 different dialogues sampled from two widely used dialogue comprehension datasets, MuTual \citep{cui-2020-mutual} and DREAM \citep{sun-2019-dream}. Then we further introduce a new dialogue comprehension model that integrates STRUDEL into a dialogue reasoning module based on graph neural networks over transformer encoder language models. Our empirical experiment results show that STRUDEL is indeed very effective in providing transformer language models with better support for reasoning and inference over challenging downstream dialogue comprehension tasks such as dialogue question answering and response prediction and improving their performance.

\section{Related Work}

\subsection{Abstractive Text Summarization}

Abstractive text summarization \citep{see-etal-2017-get, rush-etal-2015-neural,pmlr-v119-zhang20ae} is the task of generating a concise summary of a source text by paraphrasing its main content instead of simply extracting its important sentences, which is referred to as extractive summarization. Recently, new methods of abstractive summarization started to take the structure of the source document into consideration when generating a single free-form abstractive summarization. For example, \citet{balachandran-etal-2021-structsum} proposed a new summarization framework based on document-level structure induction that incorporates latent and explicit dependencies across sentences in the source document into summarization models. And \citet{wu2021bass} presented a novel framework for boosting abstractive summarization called BASS that is based on a unified semantic graph and a graph-based encoder-decoder model.

\subsection{Abstractive Dialogue Summarization}

Abstractive dialogue summarization \citep{chen-2021-dialogsum, gliwa-etal-2019-samsum,fabbri-etal-2021-convosumm,zhu-etal-2021-mediasum,zhong-etal-2021-qmsum} is a particular type of abstractive text summarization task that focuses on the abstractive summarization of human dialogues and conversations. Recently, there has been a lot of work that tries to improve the quality and faithfulness of model-generated abstractive dialogue summaries. For example, \citet{feng2020b} proposed to use heterogeneous graph networks to incorporate commonsense knowledge into abstractive dialogue summarization models. And \citet{tang-etal-2022-confit} proposed to improve the quality and faithfulness of abstractive dialogue summarization through a new training strategy based on linguistically-informed contrastive fine-tuning.

\subsection{Dialogue Comprehension and Understanding}

There have also been many advances in multi-turn dialogue comprehension and understanding \citep{zhang-2021-advances} in recent years. For example, \citet{liu2020} showed that explicitly modeling utterance-aware and speaker-aware representations can boost the dialogue comprehension performance of pre-trained language models. \citet{ouyang2020} showed that segmenting a dialogue into elementary discourse units (EDUs) and then use a graph modeling framework to model their relationships can help language models to better understand the dialogue's innate structure. \citet{zhang-zhao-2021-structural} proposed a new structural pre-training method for dialogue comprehension that captures dialogue-exclusive features by adding utterance order restoration and sentence backbone regularization into the language modeling objectives. And neural-retrieval-in-the-loop architectures have been shown to reduce hallucination in conversation models \citep{shuster2021}.

\section{Structured Dialogue Summarization}

In this section we introduce our proposed new task of structured dialogue summarization, give its detailed definition and demonstrate its features through a concrete example.

\subsection{Definition of Structured Dialogue Summarization}

\label{subsec:strudel_definition}

We define \textit{Structured Dialogue Summarization} (STRUDEL) as the task of generating a systematic and abstractive multi-entry dialogue summarization organized in a structured form that represents a comprehensive multi-aspect understanding and interpretation of a dialogue's content.

A complete STRUDEL summarization of a dialogue\footnote{In this paper we focus on the structured dialogue summarization of two-speaker dialogues, which are the most commonly seen type of dialogues in dialogue datasets and real applications. We leave the extension of STRUDEL to multi-speaker dialogues to future work (see Section \ref{sec:limitations}).} contains a set of 7 different STRUDEL entries, which are each defined as follows:

\begin{enumerate}[(a)] 
  
  \item \textbf{Relationship} - the relationship between the two speakers of the dialogue.
  
  \item \textbf{Purpose/Theme} - the main purpose or theme for which the dialogue is made between the two speakers.
  
  \item \textbf{Task/Intention}$_{\mathcal{S}_1}$ - the main task or intention that the first speaker would like to achieve in the dialogue.
  
  \item \textbf{Task/Intention}$_{\mathcal{S}_2}$ - the main task or intention that the second speaker would like to achieve in the dialogue.
  
  \item \textbf{Problem/Disagreement} - the most important problem or disagreement that the two speakers need to solve in the dialogue.
  
  \item \textbf{Solution} - the solution that the two speakers reach for the most important problem or disagreement in the dialogue.
  
  \item \textbf{Conclusion/Agreement} - the final conclusion or agreement that the two speakers reach in the dialogue.
  
\end{enumerate}

In an actual STRUDEL summarization of a dialogue, the content of each of the above 7 STRUDEL entries will either be a short text abstractively summarizing the specific aspect of the dialogue as indicated by that STRUDEL entry's definition, or be \texttt{`N/A'} indicating that the entry can't be inferred from or is not mentioned in the current dialogue.

\subsection{Example of Structured Dialogue Summarization}

\begin{figure}[t]
  \centering
  \includegraphics[width = \columnwidth]{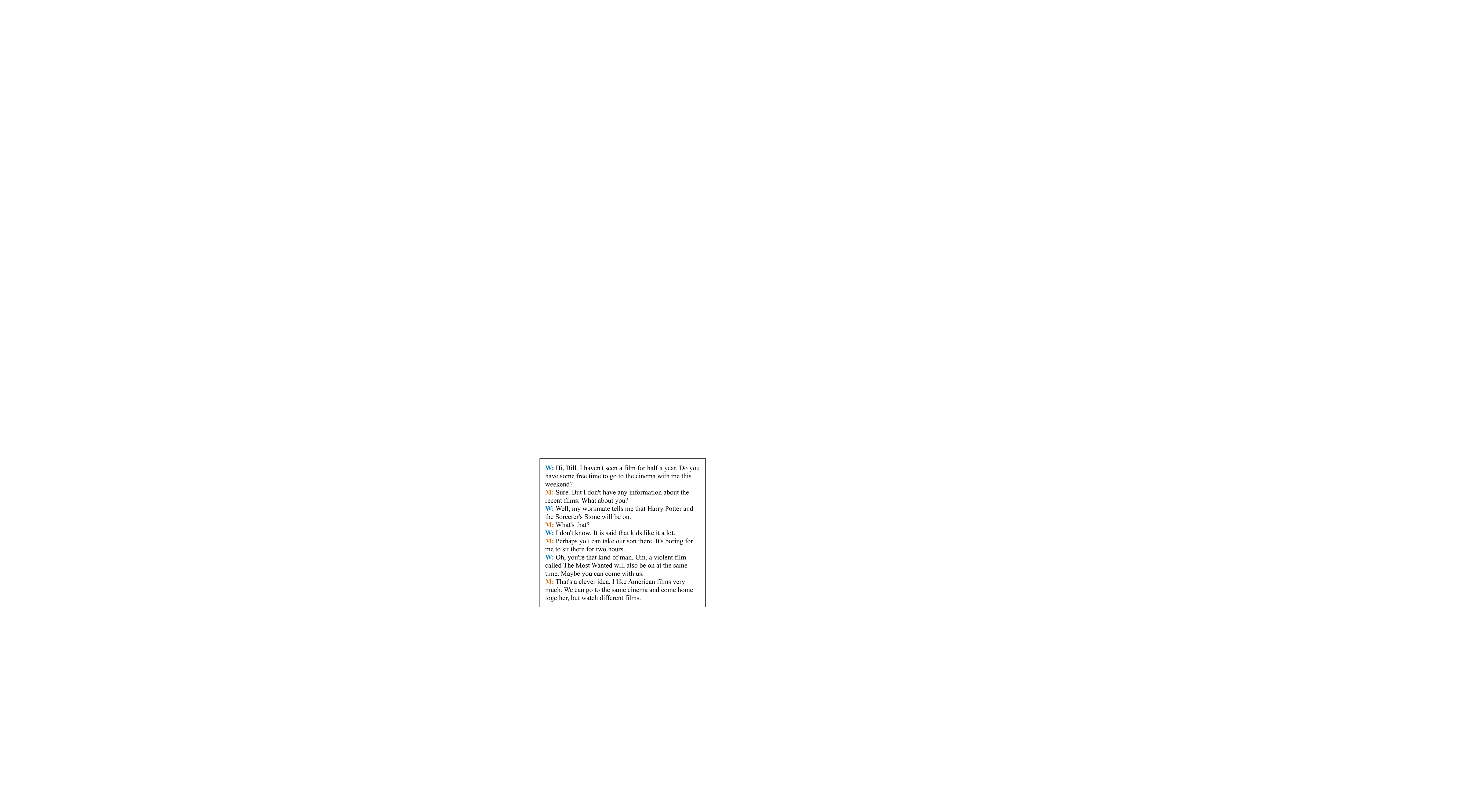}
  \caption{An example dialogue from the DREAM dataset \citep{sun-2019-dream}.}
  \label{fig:example_dialogue}
\end{figure}

Here we use a concrete example to demonstrate structured dialogue summarization of a dialogue. Figure~\ref{fig:example_dialogue} shows an example dialogue from the DREAM dataset \citep{sun-2019-dream}. For this dialogue, its structured dialogue summarization is:

\quad

\textbf{Relationship}: \textit{``Wife and husband.''}
  
\textbf{Purpose/Theme}: \textit{``Go to the cinema this weekend.''}
  
\textbf{Task/Intention}$_{\mathcal{S}_1}$: \textit{``Pick a movie to watch.''}
  
\textbf{Task/Intention}$_{\mathcal{S}_2}$: \textit{``Pick a movie to watch.''}
  
\textbf{Problem/Disagreement}: \textit{``It's boring for Bill to watch the film Happy Potter and the Sorcerer's Stone.''}
  
\textbf{Solution}: \textit{``Bill will watch another film called the Most Wanted.''}
  
\textbf{Conclusion/Agreement}: \textit{``Go to the cinema and come home together, but watch different films.''}

\quad

This same example also appears in the DIALOGSUM dataset \citep{chen-2021-dialogsum}, which is a dataset for traditional abstractive dialogue summarization. In contrast, this dialogue's traditional abstractive summarization annotated in the DIALOGSUM dataset is the following:

\quad

\textit{``Person1 invites Bill to go to the cinema together this weekend. Person1 hears the Harry Potter movie would be on but Person2 likes the violent film.''}

\quad

From this comparison between the traditional free-form abstractive dialogue summarization and our proposed structured dialogue summarization, we can clearly see that the STRUDEL summarization includes more important aspects about the dialogue and tells a more comprehensive and informative story compared to the traditional free-form abstractive dialogue summarization.

\section{Human Annotations of STRUDEL}

Our proposed new task of Structured Dialogue Summarization (STRUDEL) opens up a gateway for language models to observe, imitate, and learn from the structured human mental process of systematic dialogue understanding. But in order to actually infuse these valuable human-guided structural priors regarding dialogue understanding into language models through the task of STRUDEL, we first need to collect high-quality supervision information from the empirical human demonstration of performing the STRUDEL task. Therefore, for this purpose, we collect a set of human annotations of STRUDEL over 400 dialogues sampled from two widely used dialogue comprehension datasets - the MuTual dataset \citep{cui-2020-mutual} for dialogue response prediction and the DREAM dataset \citep{sun-2019-dream} for dialogue question answering. In our collection of STRUDEL human annotations, each sampled dialogue is manually annotated with its complete set of STRUDEL summarization with all 7 STRUDEL entries (can contain \texttt{`N/A'}) by a human annotator following the annotation protocols (see Section \ref{sec:annotation_protocols}).

\subsection{Datasets}
\label{subsec:datasets}

The two dialogue comprehension datasets that we used for the human annotations of STRUDEL are:

\subsubsection{MuTual}

\begin{table}[t]
	\centering\small
	 \resizebox{\linewidth}{!}
    {\setlength{\tabcolsep}{7.5pt}
		\begin{tabular}{ccccc}
			\toprule
			
			STRUDEL & \multicolumn{2}{c}{\textbf{DREAM}} & \multicolumn{2}{c}{\textbf{MuTual}}  \\
      \cmidrule(r){2-3} \cmidrule(r){4-5}
			Entry Name &\textbf{\%} & \textbf{Avg Len} & \textbf{\%} & \textbf{Avg Len} \\
      \midrule
			\midrule
      Relationship   &  72\% & 1.92& 54.5\%& 1.69\\
      \midrule
      \makecell{Purpose/\\Theme}   &  100\% & 7.80& 100 \%& 6.55\\
      \midrule
      \makecell{Task/\\Intention$_{\mathcal{S}_1}$}   &  99.5\% & 7.36& 99.5\%& 7.08\\
      \midrule
      \makecell{Task/\\Intention$_{\mathcal{S}_2}$}   & 99\% & 7.10&97\% &  6.51\\
      \midrule
      \makecell{Problem/Dis\\agreement}   &  94.5\% & 10.76& 91.5\%& 9.97\\
      \midrule
      Solution   &  92.5\% & 12.33& 90.5\%& 11.25\\
      \midrule
      \makecell{Conclusion/\\Agreement}   &  90\%& 14.83& 88\%& 15.74\\
			\bottomrule
		\end{tabular}
	}
	\caption{\label{tab:annotation_statistics}  Statistics of our collected STRUDEL human annotations over the DREAM dataset and the MuTual dataset. '\textbf{\%}' denotes frequency of appearance (i.e. not annotated with \texttt{`N/A'}) in percentage, and '\textbf{Avg Len}' denotes average length of each STRUDEL summarization entry as measured by number of words.}
\end{table}

MuTual \citep{cui-2020-mutual} is a popular recently proposed multi-turn dialogue reasoning dataset in the form of dialogue response prediction. All dialogue corpora in the MuTual dataset are modified from Chinese high school English listening comprehension test data, where students are expected to select the best answer from three candidate options, given a multi-turn dialogue and a question. The authors asked human annotators to rewrite the question and answer candidates as response candidates to fit in the test scenario of dialogue response prediction. MuTual has a total of 8860 expert-designed challenge questions, almost all of which require reasoning.

\subsubsection{DREAM}

DREAM \citep{sun-2019-dream} is the first multiple-choice reading comprehension dataset on dialogues. It is collected from English comprehension examinations designed by human experts and contains 10197 multiple-choice questions for 6444 different dialogues. DREAM presents a challenging in-depth, multi-turn, and multi-party dialogue understanding task because of its features of being mostly non-extractive, requiring reasoning beyond single sentences, and involving commonsense knowledge. 

\begin{figure*}[t]
  \centering
  \includegraphics[width = \textwidth]{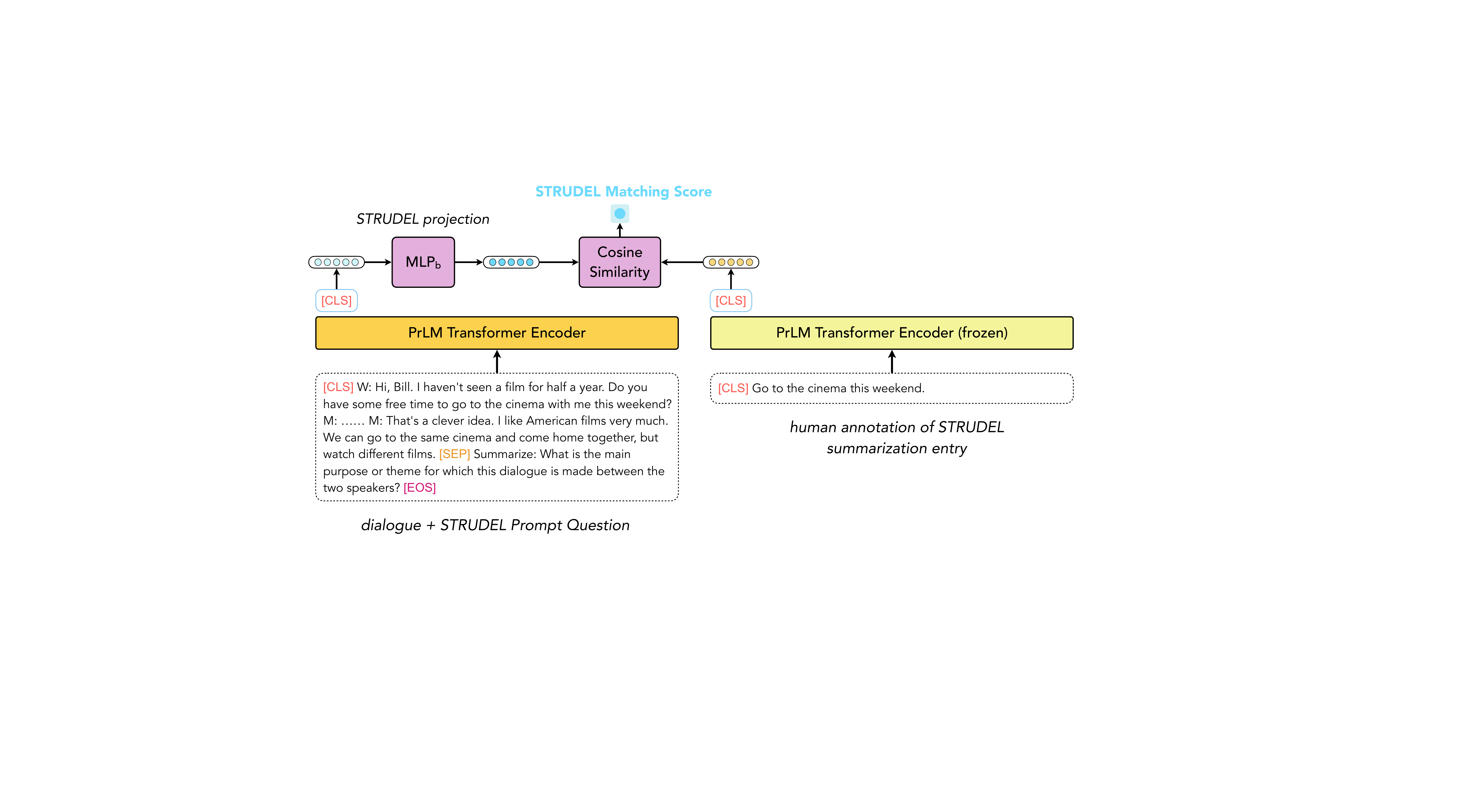}
  \caption{The modeling pipeline that trains a transformer encoder to learn to generate vector embeddings of STRUDEL entries that match their corresponding human annotations.}
  \label{fig:strudel_matching}
\end{figure*}

\subsection{Annotation Protocols}
\label{sec:annotation_protocols}

The two major annotation protocols that we prescribed to the annotators during the STRUDEL human annotation process are:

\begin{enumerate}
  \item When writing each STRUDEL summarization entry for a dialogue, please be informative, concise, faithful, and to the point.
  \item When you think a certain STRUDEL entry can’t be inferred from the dialogue or is not mentioned in the dialogue at all or doesn’t apply to the current dialogue, please write \texttt{`N/A'} for that STRUDEL entry in your annotation.
\end{enumerate}

\subsection{Annotation Statistics}

The statistics of our collected human annotations of STRUDEL are reported in Table \ref{tab:annotation_statistics}. From these statistics, we can see that the majority of STRUDEL entries have high appearance frequency across dialogues, and on average the annotated STRUDEL entries are relatively concise.

\section{Modeling Approach}

In this section, we describe our main modeling approach that uses Structured Dialogue Summarization (STRUDEL) to improve pre-trained language models' abilities of dialogue comprehension.

\subsection{STRUDEL as a Meta-Model}

As we can see from the definition in Section \ref{subsec:strudel_definition}, Structured Dialogue Summarization (STRUDEL) is a generic task that can be generally applied to any dialogue. Therefore, STRUDEL can be viewed as an important upstream auxiliary NLU task and can be used to train language models to better understand dialogues in a structured and systematic way before they were further fine-tuned over specific downstream dialogue comprehension tasks.

As a result, based on our definition of STRUDEL, we further propose a new modeling framework of STRUDEL dialogue comprehension, in which STRUDEL can be viewed as a meta-model that can be smoothly integrated into and used on top of a wide range of different large-scale pre-trained transformer encoder models for dialogue understanding. Figure \ref{fig:meta_model} provides a conceptual illustration of this relationship between STRUDEL and pre-trained language models. Below we discuss each of the different components of our STRUDEL dialogue comprehension modeling framework in detail.

\subsection{STRUDEL Prompt Questions}

\label{subsec:prompt_questions}

We first design a prompt question for each STRUDEL summarization entry, which will be used to query a pre-trained language model to generate a vector embedding of that STRUDEL entry for a dialogue. For each STRUDEL summarization entry defined in Section \ref{subsec:strudel_definition}, we add the common prefix \texttt{`Summarize: what is '} to its definition sentence and replace the \texttt{`.'} at the end with \texttt{`?'} to form its corresponding STRUDEL prompt question. For example, for STRUDEL entry (a), the \textit{relationship} entry, its definition sentence is \texttt{`the relationship between the two speakers of the dialogue.'}, and its corresponding STRUDEL prompt question is \texttt{`Summarize: what is the relationship between the two speakers of the dialogue?'}

\subsection{Learning to Generate STRUDEL Embeddings}

\label{subsec:learning_to_generate_STRUDEL_embeddings}

In our STRUDEL dialogue comprehension modeling framework, we choose to train transformer encoder language models to learn to generate semantic vector embeddings of the contents of STRUDEL entries instead of the actual textual outputs of the STRUDEL entries in the form of token sequences. We make this design choice mainly for two reasons: (1) the form of vector embeddings makes it easier to quantitatively compare model-generated structured dialogue summarizations with their corresponding human annotations (e.g. by calculating cosine similarities in the vector space); (2) vector embeddings of STRUDEL can also be smoothly integrated into a reasoning module based on graph neural networks for running inference over dialogue comprehension tasks.

Now we describe the procedure to train a pre-trained transformer encoder language model to learn to generate STRUDEL embeddings under the supervision of STRUDEL human annotations. Given a dialogue input sequence $D$ and a pre-trained transformer encoder language model $\mathcal{T}$ for computing deep contextualized representations of textual sequences, such as BERT \citep{devlin-etal-2019-bert} and RoBERTa \citep{liu2019roberta}, for an entry $\mathcal{E}$ of the STRUDEL summarization, we first concatenate $D$ with the STRUDEL prompt question $Q_{\mathcal{E}}$ for the STRUDEL entry $\mathcal{E}$ (as defined in Section \ref{subsec:prompt_questions}) together to form a query sequence $\{\texttt{[CLS]} D \texttt{[SEP]} Q_{\mathcal{E}} \texttt{[EOS]} \}$, and then feed this query sequence into the transformer encoder $\mathcal{T}$ to compute its contextualized representation. Let $H^{\mathcal{E}}$ be the last layer of hidden state vectors computed from this transformer encoder $\mathcal{T}$, then we have:
$$
  H^{\mathcal{E}} = \mathcal{T}\Big(\{\texttt{[CLS]}  D  \texttt{[SEP]}  Q_{\mathcal{E}}  \texttt{[EOS]} \}\Big)
$$

Let $h_{\texttt{[CLS]}}^{\mathcal{E}}$ denote the last-layer hidden state vector of the \texttt{[CLS]} token in $H^{\mathcal{E}}$, then we apply a dedicated multi-layer perceptron MLP$^{\mathcal{E}}$ on top of $h_{\texttt{[CLS]}}^{\mathcal{E}}$ to project it onto a same-dimensional vector space to obtain our final vector embedding of the STRUDEL entry $\mathcal{E}$.

Now let $A^{\mathcal{E}}$ denote the human-annotated ground-truth summarization for STRUDEL entry $\mathcal{E}$. Then we use a frozen version of the same transformer encoder, denoted as $\tilde{\mathcal{T}}$, to encode this human annotation as: 
$$
  \tilde{H}^{\mathcal{E}} = \tilde{\mathcal{T}}\Big(\{\texttt{[CLS]} A^{\mathcal{E}}  \texttt{[EOS]} \} \Big)
$$

Let $\tilde{h}_{\texttt{[CLS]}}^{\mathcal{E}}$ denote the last-layer hidden state vector of the \texttt{[CLS]} token in $\tilde{H}^{\mathcal{E}}$, then we can compute the semantic matching score between the transformer model's generated vector embedding for STRUDEL entry $\mathcal{E}$ and its corresponding human annotation as their cosine similarity value: $\text{cos}\Big(\text{MLP}^{\mathcal{E}}(h_{\texttt{[CLS]}}^{\mathcal{E}}), \; \tilde{h}_{\texttt{[CLS]}}^{\mathcal{E}}\Big)$. Therefore, the objective function for optimizing the transformer encoder model $\mathcal{T}$ to generate STRUDEL summaries that match human annotations can be formulated as:

\begin{align}
  \mathbb{L}_{\text{SM}} = - \sum_{\mathcal{E}} \text{cos}\Big(\text{MLP}^{\mathcal{E}}(h_{\texttt{[CLS]}}^{\mathcal{E}}), \; \tilde{h}_{\texttt{[CLS]}}^{\mathcal{E}}\Big) \label{eq:semantic_matching}
\end{align}

See Figure \ref{fig:strudel_matching} for an illustration of this modeling pipeline. 

\begin{figure}[t]
  \centering
  \includegraphics[width = \columnwidth]{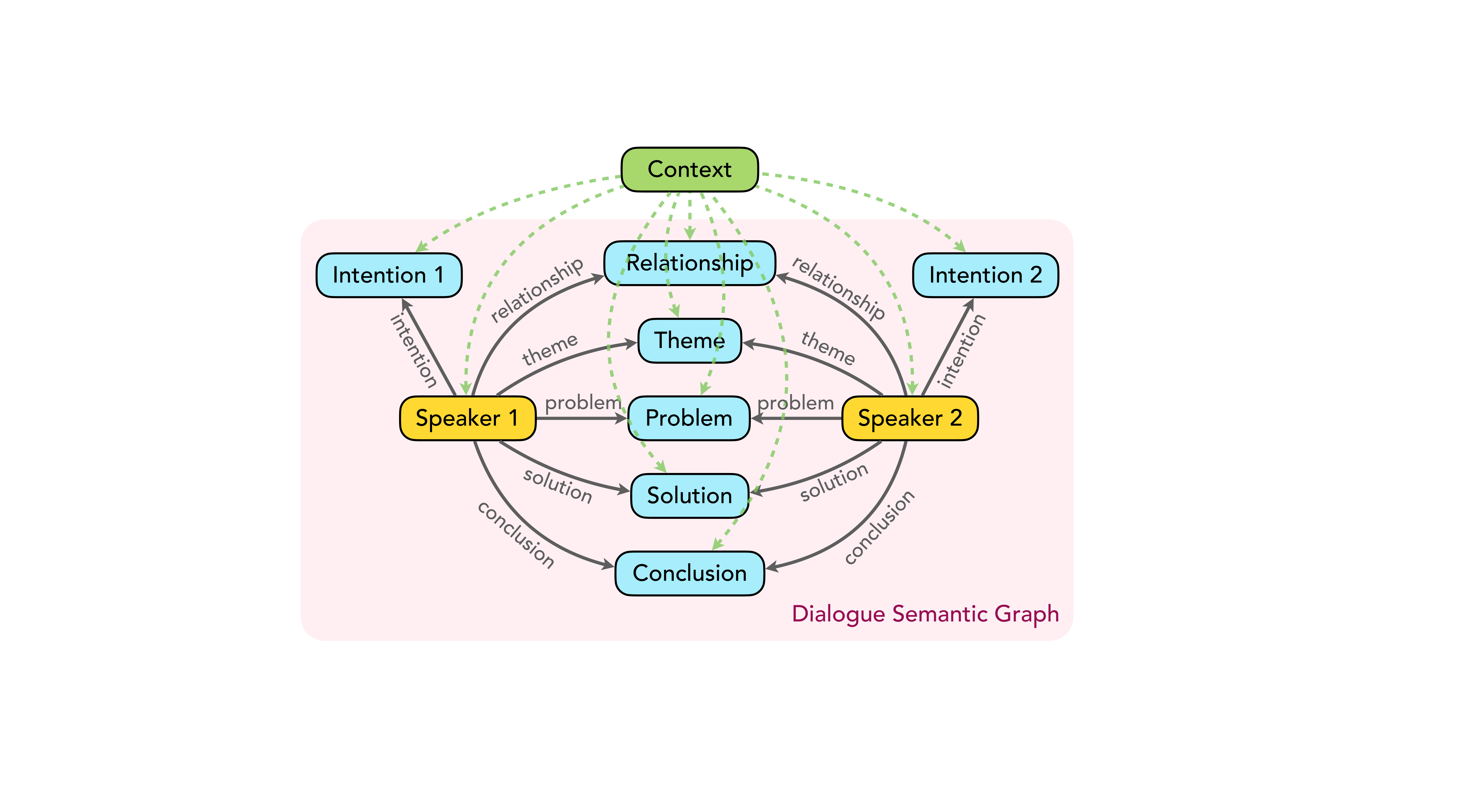}
  \caption{The dialogue semantic graph of STRUDEL embeddings and the context node on top of the graph. The speaker nodes are depicted in color \textcolor{yellow1}{yellow} and the STRUDEL embedding nodes are depicted in color \textcolor{blue1}{blue}.}
  \label{fig:dialogue_semantic_graph}
\end{figure}

\begin{figure*}[t]
  \centering
  \includegraphics[width = \textwidth]{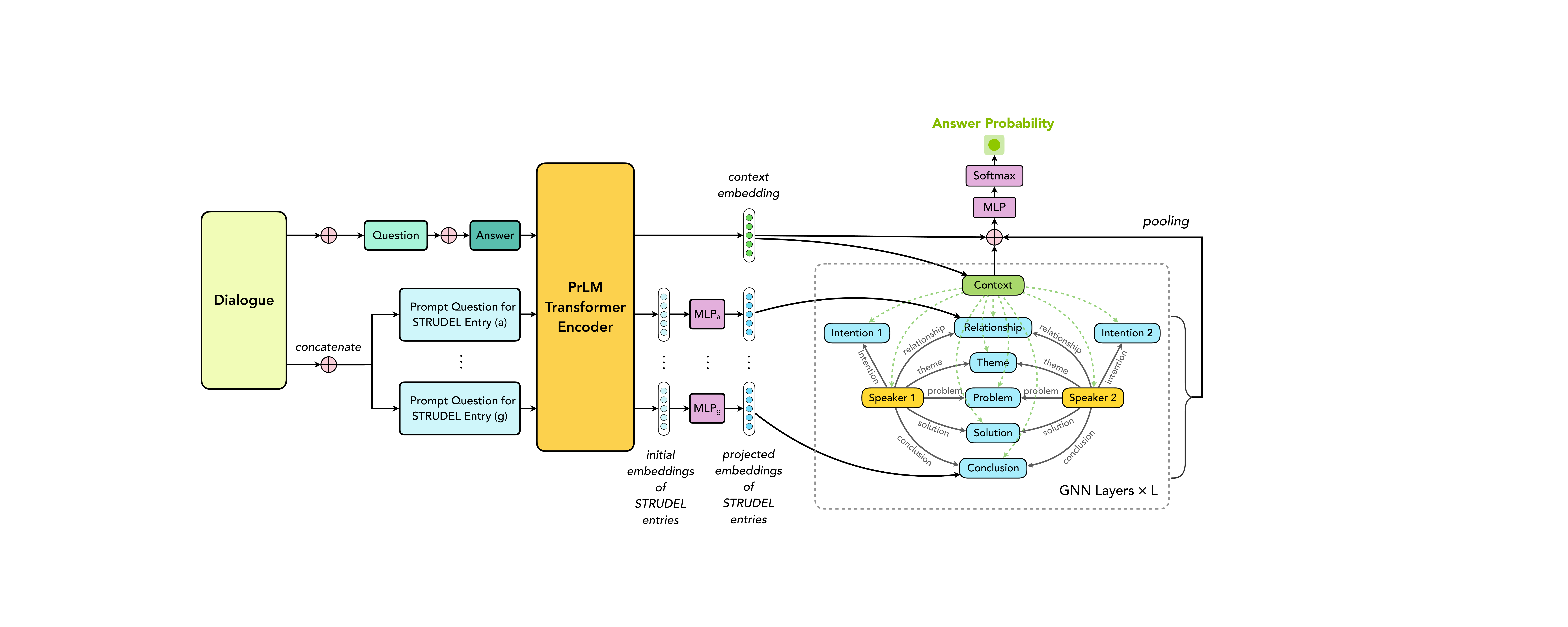}
  \caption{The overall model architecture of our STRUDEL dialogue comprehension modeling framework.}
  \label{fig:model_architecture}
\end{figure*}

\subsection{Dialogue Semantic Graph}

In order to systematically organize all the information contained in different STRUDEL summary entries into a structured representation that is amenable to collective reasoning and inference, we propose to construct a graph representation of STRUDEL embeddings called \textbf{dialogue semantic graph}. The dialogue semantic graph consists of two types of nodes: 2 speaker nodes (one for the first speaker and the other one for the second speaker) and 7 STRUDEL embedding nodes (one for each of the 7 STRUDEL summary entries). It also has 6 different types of relational edges, each corresponding to one of the following 6 different relations from speakers to STRUDEL summaries: `\textit{has\_relationship}', `\textit{shares\_theme}', `\textit{with\_intention}', `\textit{faces\_problem}', `\textit{finds\_solution}', and `\textit{reaches\_conclusion}'. The structure of the dialogue semantic graph is depicted in Figure \ref{fig:dialogue_semantic_graph}.

\subsection{STRUDEL for Dialogue Comprehension}

After a transformer encoder language model learns to generate embeddings of structured dialogue summarization, we need to design a modeling framework to employ these generated STRUDEL embeddings to improve the model's dialogue comprehension abilities. In this paper we focus on two important types of dialogue comprehension tasks - dialogue question answering and dialogue response prediction \citep{zhang-2021-advances}. Since all the generated STRUDEL embeddings can be organized into a dialogue semantic graph, we can build a graph neural network (GNN) module on top of the dialogue semantic graph to perform reasoning for different dialogue comprehension tasks. In this paper, we adopt the QA-GNN \citep{yasunaga-etal-2021-qa} architecture as our GNN reasoning module to perform context-aware structured reasoning over generated STRUDEL embeddings.

More specifically, given a dialogue input sequence $D$, a question $Q$, a candidate answer $A$ (for dialogue response prediction tasks, $Q$ will be empty and $A$ will be a candidate response) and a transformer encoder language model $\mathcal{T}$, we first use the procedure described in Section \ref{subsec:learning_to_generate_STRUDEL_embeddings} to compute the STRUDEL embedding vector $v^{\mathcal{E}}$ for each STRUDEL entry $\mathcal{E}$ as: $v^{\mathcal{E}}$ = $\text{MLP}^{\mathcal{E}}(h_{\texttt{[CLS]}}^{\mathcal{E}})$. Then we form a context sequence $\{\texttt{[CLS]} D \texttt{[SEP]} Q \texttt{[SEP]} A \texttt{[EOS]} \}$ and feed it into the same transformer encoder $\mathcal{T}$ to compute its contextualized representation:
$$
  H = \mathcal{T}\Big(\{\texttt{[CLS]} D \texttt{[SEP]} Q \texttt{[SEP]} A \texttt{[EOS]} \} \Big).
$$

Let $h_{\texttt{[CLS]}}$ denote the last-layer hidden state vector of the \texttt{[CLS]} token in $H$, then we take $h_{\texttt{[CLS]}}$ as the context embedding vector $v^{\mathcal{C}}$ that encodes the original dialogue, the question and the current answer candidate (for dialogue response prediction tasks it encodes the original dialogue and the current response option): $v^{\mathcal{C}} = h_{\texttt{[CLS]}}$. And additionally, for the two speakers $\mathcal{S}_1$ and $\mathcal{S}_2$ of the dialogue, we use the frozen transformer encoder $\tilde{\mathcal{T}}$ to compute the following two contextualized representations:

\vspace{-1em}

\begin{equation*}
  \resizebox{.98\hsize}{!}{$H^{\mathcal{S}_1} = \tilde{\mathcal{T}}\Big(\{\text{`\textit{The first speaker of this dialogue.}'}\} \Big)$,}
\end{equation*}

\vspace{-2em}

\begin{equation*}
  \resizebox{.98\hsize}{!}{$H^{\mathcal{S}_2} = \tilde{\mathcal{T}}\Big(\{\text{`\textit{The second speaker of this dialogue.}'}\} \Big)$,}
\end{equation*}

and assign the embedding vectors of the two speakers to be the last-layer hidden state vectors of the corresponding \texttt{[CLS]} tokens: $v^{\mathcal{S}_1} = h^{\mathcal{S}_1}_{\texttt{[CLS]}}$ and $v^{\mathcal{S}_2} = h^{\mathcal{S}_2}_{\texttt{[CLS]}}$.

\begin{table*}[t]
  {
      \centering\small
      \resizebox{0.75\linewidth}{!}
      { \setlength{\tabcolsep}{8pt}
          \begin{tabular}{l  l  l  l  l }
              \toprule 
              & \multicolumn{3}{c}{MuTual} & \multicolumn{1}{c}{DREAM} \\
              \cmidrule(r){2-4} \cmidrule(r){5-5}
              \textbf{Model} & $\textbf{R}_{4}$@1 & $\textbf{R}_{4}$@2 & MRR & Accuracy  \\
              \midrule \midrule
               BERT-large \citep{devlin-etal-2019-bert} & 0.662  & 0.858 & 0.803 & 0.636 \\
              \midrule
               BERT-large + STRUDEL & \textbf{0.727} & \textbf{0.881} & \textbf{0.839} & \textbf{0.668}   \\
              \midrule
               RoBERTa-large \citep{liu2019roberta} & 0.786  & 0.931 & 0.879 & 0.821 \\ 
              \midrule
               RoBERTa-large + STRUDEL & \textbf{0.887}  & \textbf{0.969} & \textbf{0.937} & \textbf{0.839} \\
                 \bottomrule
          \end{tabular}
      }
      \caption{\label{tab:result} Our experiment results on the MuTual dataset \citep{cui-2020-mutual} and the DREAM dataset \citep{sun-2019-dream}. Following \citet{cui-2020-mutual}, we report recall at position 1 in 4 candidates ($\textbf{R}_{4}$@1), recall at position 2 in 4 candidates ($\textbf{R}_{4}$@2) and Mean Reciprocal Rank (MRR) for the models' performance on the MuTual dataset. The higher value in each comparison pair is highlighted in \textbf{bold}.
    }
  }
\end{table*}

Following the architecture of QA-GNN \citep{yasunaga-etal-2021-qa}, to build our GNN reasoning module over STRUDEL embeddings, we add a context node on top of the dialogue semantic graph and add 9 edges between the context node and all the 9 nodes in the dialogue semantic graph to form a joint graph for dialogue reasoning (see Figure \ref{fig:dialogue_semantic_graph}). Then we construct an $L$-layer graph attention network (GAT) \citep{velickovic2018graph} using this joint graph and integrate it with the central transformer encoder language model $\mathcal{T}$ to perform reasoning over dialogue comprehension tasks. More specifically, we first add dimension projection layers to project the context embedding vector $v^{\mathcal{C}}$, the two speaker embedding vectors $v^{\mathcal{S}_1}$ and $v^{\mathcal{S}_2}$, and all the STRUDEL embedding vectors $v^{\mathcal{E}}$ to the dimensionality of the QA-GNN node embeddings. Then we use the dimension-projected context embedding vector $\hat{v}^{\mathcal{C}}$ to initialize the context node, use the two dimension-projected speaker embedding vectors $\hat{v}^{\mathcal{S}_1}$ and $\hat{v}^{\mathcal{S}_2}$ to initialize the two speaker nodes, and use all the dimension-projected STRUDEL embedding vectors $\hat{v}^{\mathcal{E}}$ to initialize the corresponding STRUDEL entry nodes in our joint graph.

And similar to QA-GNN, after $L$-layers of GNN message passing over the joint graph, we concatenate the context embedding vector, the last-layer representation vector of the context node and the pooling of the last-layer representation vectors of all the nodes in the dialogue semantic graph together into a single vector and feed it into another multi-layer perceptron to obtain the final probability score for $(D, Q, A)$. We then apply the softmax function over the probability scores of all the different answer (or response) options to obtain their inferred probability distribution as:

\vspace{-2em}

\begin{equation*}
  \resizebox{0.98\hsize}{!}{$\mathbb{P}(A \mid D, Q) = \text{Softmax}\Big(\text{QA-GNN}(v^{\mathcal{C}}, v^{\mathcal{S}_1}, v^{\mathcal{S}_1}, \{ v^{\mathcal{E}}  \}   )\Big)$}
\end{equation*}

\vspace{-0.5em}

Let $a^{*}$ denote the correct answer (or response) provided by the training label, then the objective function that we use to train the transformer encoder language model $\mathcal{T}$ to use STRUDEL embeddings to perform dialogue question answering (or response prediction) can be formulated as the cross-entropy loss:
\begin{align}
 \mathbb{L}_{\text{CE}} = - \log \Big( \mathbb{P}(A = a^{*} \mid D, Q) \Big) \label{eq:cross_entropy}
\end{align}

See Figure \ref{fig:model_architecture} for an illustration of the overall model architecture for STRUDEL dialogue comprehension described above.

\subsection{Model Training}

\subsubsection{Multi-Task Post-Training}

During the training of our STRUDEL dialogue comprehension model, we first adopt a multi-task learning strategy to train our model to learn to generate accurate STRUDEL embeddings and to infer the correct choices for dialogue question answering and response prediction tasks based on its generated STRUDEL embeddings at the same time. This multi-task post-training process uses distinct but complementary tasks to challenge the model to learn structured and meaningful representations of dialogue semantics that are widely generalizable to different dialogue comprehension tasks. To do this, we define our objective function to be a weighted sum of the average of the semantic matching loss (defined in Equation \ref{eq:semantic_matching}) across all the human-annotated dialogues and the average of the cross-entropy loss (defined in Equation \ref{eq:cross_entropy}) across all the dialogue comprehension examples:

\begin{align}
  \mathbb{L} = \frac{1}{N_{\textit{ha}}} \sum^{N_{\textit{ha}}}_{i = 1} \Big(\alpha_1 \mathbb{L}^{i}_{\text{SM}} \Big) + \frac{1}{N} \sum^{N}_{j = 1} \Big(\alpha_2 \mathbb{L}^{j}_{\text{CE}}\Big) \label{eq:overall_loss_function}
\end{align}

where $N$ is the total number of dialogue comprehension examples, $N_{\textit{ha}}$ is the total number of dialogues annotated with STRUDEL summaries in the human annotation process, and $\alpha_1$ and $\alpha_2$ are the hyperparameters that correspond to the weights in the weighted sum.

\subsubsection{Single-Task Fine-Tuning}

After our transformer-based STRUDEL dialogue comprehension model has been post-trained using the objective function defined in Equation \ref{eq:overall_loss_function}, we take the post-trained model checkpoint and continue to fine-tune the model over individual dialogue comprehension tasks in order to fully maximize its performance on each of the tasks. 

\section{Experiments}

\subsection{Transformer Encoder Models}

In our experiment, we use two widely-used transformer encoder language models - BERT \citep{devlin-etal-2019-bert} and RoBERTa \citep{liu2019roberta} - as the backbone transformer encoder in our STRUDEL dialogue comprehension modeling framework.

\subsection{Dialogue Comprehension Tasks}

In our experiment, we test our STRUDEL dialogue comprehension model on two important and representative dialogue comprehension tasks - dialogue question answering and dialogue response prediction. We use the MuTual dataset \citep{cui-2020-mutual} and the DREAM dataset \citep{sun-2019-dream} introduced in Section \ref{subsec:datasets} to train and test our model.

\subsection{Results}

The results of our experiments are shown in Table \ref{tab:result}. As we can see from the table, the performance scores of our STRUDEL dialogue comprehension models on both the dialogue response prediction task (over the MuTual dataset) and the dialogue question answering task (over the DREAM dataset) are all consistently higher than their corresponding backbone transformer encoder models alone. This demonstrates that our proposed task of Structured Dialogue Summarization (STRUDEL) and our proposed STRUDEL dialogue comprehension modeling framework can indeed help transformer language models learn to better perform dialogue comprehension tasks.

\section{Conclusion}

In this paper, we presented STRUDEL (STRUctured DiaLoguE Summarization) - a novel type of dialogue summarization task that can help pre-trained language models to better understand dialogues and improve their performance on important dialogue comprehension tasks. In contrast to the traditional free-form abstractive summarization task for dialogues, STRUDEL provides a more comprehensive digest over multiple important aspects of a dialogue and has the advantage of being more focused, specific, and instructive for dialogue comprehension models to learn from. In addition, we also introduced a new STRUDEL dialogue comprehension modeling framework that integrates STRUDEL into a GNN-based dialogue reasoning module over transformer encoder language models to improve their dialogue comprehension abilities. Our empirical experiments on the tasks of dialogue question answering and dialogue response prediction showed that our STRUDEL dialogue comprehension modeling framework can significantly improve the dialogue comprehension performance of transformer encoder language models.

\section{Limitations}
\label{sec:limitations}

There are two limitations of our work discussed in this paper:

\begin{enumerate}
 \item Our paper mainly focuses on designing the structured dialogue summarization task for two-speaker dialogues, which is the majority of multi-turn dialogues that are most commonly seen in dialogue datasets and real applications. In the future, we plan to further extend our STRUDEL framework to also accommodate multi-speaker dialogues among more than two speakers.
 \item Our approach hasn't included explicit knowledge reasoning components yet, which are also important for language models to accurately generate structured dialogue summarizations and to perform dialogue comprehension tasks. In future work, we plan to integrate a knowledge reasoning module into our STRUDEL dialogue comprehension modeling framework in order to further improve its performance.
\end{enumerate}

\section*{Acknowledgment}

We would like to thank Meta AI for their generous support of our work.

\bibliography{anthology,custom}

\begin{thebibliography}{23}
\expandafter\ifx\csname natexlab\endcsname\relax\def\natexlab#1{#1}\fi

\bibitem[{Balachandran et~al.(2021)Balachandran, Pagnoni, Lee, Rajagopal,
  Carbonell, and Tsvetkov}]{balachandran-etal-2021-structsum}
Vidhisha Balachandran, Artidoro Pagnoni, Jay~Yoon Lee, Dheeraj Rajagopal, Jaime
  Carbonell, and Yulia Tsvetkov. 2021.
\newblock \href {https://doi.org/10.18653/v1/2021.eacl-main.220}
  {{S}truct{S}um: Summarization via structured representations}.
\newblock In \emph{Proceedings of the 16th Conference of the European Chapter
  of the Association for Computational Linguistics: Main Volume}, pages
  2575--2585, Online. Association for Computational Linguistics.

\bibitem[{Chen et~al.(2021)Chen, Liu, Chen, and Zhang}]{chen-2021-dialogsum}
Yulong Chen, Yang Liu, Liang Chen, and Yue Zhang. 2021.
\newblock \href {https://doi.org/10.18653/v1/2021.findings-acl.449}
  {{D}ialog{S}um: {A} real-life scenario dialogue summarization dataset}.
\newblock In \emph{Findings of the Association for Computational Linguistics:
  ACL-IJCNLP 2021}, pages 5062--5074, Online. Association for Computational
  Linguistics.

\bibitem[{Cui et~al.(2020)Cui, Wu, Liu, Zhang, and Zhou}]{cui-2020-mutual}
Leyang Cui, Yu~Wu, Shujie Liu, Yue Zhang, and Ming Zhou. 2020.
\newblock \href {https://doi.org/10.18653/v1/2020.acl-main.130} {{M}u{T}ual: A
  dataset for multi-turn dialogue reasoning}.
\newblock In \emph{Proceedings of the 58th Annual Meeting of the Association
  for Computational Linguistics}, pages 1406--1416, Online. Association for
  Computational Linguistics.

\bibitem[{Devlin et~al.(2019)Devlin, Chang, Lee, and
  Toutanova}]{devlin-etal-2019-bert}
Jacob Devlin, Ming-Wei Chang, Kenton Lee, and Kristina Toutanova. 2019.
\newblock \href {https://doi.org/10.18653/v1/N19-1423} {{BERT}: Pre-training of
  deep bidirectional transformers for language understanding}.
\newblock In \emph{Proceedings of the 2019 Conference of the North {A}merican
  Chapter of the Association for Computational Linguistics: Human Language
  Technologies, Volume 1 (Long and Short Papers)}, pages 4171--4186,
  Minneapolis, Minnesota. Association for Computational Linguistics.

\bibitem[{Fabbri et~al.(2021)Fabbri, Rahman, Rizvi, Wang, Li, Mehdad, and
  Radev}]{fabbri-etal-2021-convosumm}
Alexander Fabbri, Faiaz Rahman, Imad Rizvi, Borui Wang, Haoran Li, Yashar
  Mehdad, and Dragomir Radev. 2021.
\newblock \href {https://doi.org/10.18653/v1/2021.acl-long.535} {{C}onvo{S}umm:
  Conversation summarization benchmark and improved abstractive summarization
  with argument mining}.
\newblock In \emph{Proceedings of the 59th Annual Meeting of the Association
  for Computational Linguistics and the 11th International Joint Conference on
  Natural Language Processing (Volume 1: Long Papers)}, pages 6866--6880,
  Online. Association for Computational Linguistics.

\bibitem[{Feng et~al.(2020)Feng, Feng, Qin, and Liu}]{feng2020b}
Xiachong Feng, Xiaocheng Feng, Bing Qin, and Ting Liu. 2020.
\newblock \href {http://arxiv.org/abs/2010.10044} {Incorporating commonsense
  knowledge into abstractive dialogue summarization via heterogeneous graph
  networks}.
\newblock \emph{CoRR}, abs/2010.10044.

\bibitem[{Gliwa et~al.(2019)Gliwa, Mochol, Biesek, and
  Wawer}]{gliwa-etal-2019-samsum}
Bogdan Gliwa, Iwona Mochol, Maciej Biesek, and Aleksander Wawer. 2019.
\newblock \href {https://doi.org/10.18653/v1/D19-5409} {{SAMS}um corpus: A
  human-annotated dialogue dataset for abstractive summarization}.
\newblock In \emph{Proceedings of the 2nd Workshop on New Frontiers in
  Summarization}, pages 70--79, Hong Kong, China. Association for Computational
  Linguistics.

\bibitem[{Liu et~al.(2020)Liu, Zhang, Zhao, Zhou, and Zhou}]{liu2020}
Longxiang Liu, Zhuosheng Zhang, Hai Zhao, Xi~Zhou, and Xiang Zhou. 2020.
\newblock \href {http://arxiv.org/abs/2009.06504} {Filling the gap of
  utterance-aware and speaker-aware representation for multi-turn dialogue}.
\newblock \emph{CoRR}, abs/2009.06504.

\bibitem[{Liu et~al.(2019)Liu, Ott, Goyal, Du, Joshi, Chen, Levy, Lewis,
  Zettlemoyer, and Stoyanov}]{liu2019roberta}
Yinhan Liu, Myle Ott, Naman Goyal, Jingfei Du, Mandar Joshi, Danqi Chen, Omer
  Levy, Mike Lewis, Luke Zettlemoyer, and Veselin Stoyanov. 2019.
\newblock Roberta: A robustly optimized bert pretraining approach.
\newblock \emph{arXiv preprint arXiv:1907.11692}.

\bibitem[{Ouyang et~al.(2020)Ouyang, Zhang, and Zhao}]{ouyang2020}
Siru Ouyang, Zhuosheng Zhang, and Hai Zhao. 2020.
\newblock \href {http://arxiv.org/abs/2012.14827} {Dialogue graph modeling for
  conversational machine reading}.
\newblock \emph{CoRR}, abs/2012.14827.

\bibitem[{Rush et~al.(2015)Rush, Chopra, and Weston}]{rush-etal-2015-neural}
Alexander~M. Rush, Sumit Chopra, and Jason Weston. 2015.
\newblock \href {https://doi.org/10.18653/v1/D15-1044} {A neural attention
  model for abstractive sentence summarization}.
\newblock In \emph{Proceedings of the 2015 Conference on Empirical Methods in
  Natural Language Processing}, pages 379--389, Lisbon, Portugal. Association
  for Computational Linguistics.

\bibitem[{See et~al.(2017)See, Liu, and Manning}]{see-etal-2017-get}
Abigail See, Peter~J. Liu, and Christopher~D. Manning. 2017.
\newblock \href {https://doi.org/10.18653/v1/P17-1099} {Get to the point:
  Summarization with pointer-generator networks}.
\newblock In \emph{Proceedings of the 55th Annual Meeting of the Association
  for Computational Linguistics (Volume 1: Long Papers)}, pages 1073--1083,
  Vancouver, Canada. Association for Computational Linguistics.

\bibitem[{Shuster et~al.(2021)Shuster, Poff, Chen, Kiela, and
  Weston}]{shuster2021}
Kurt Shuster, Spencer Poff, Moya Chen, Douwe Kiela, and Jason Weston. 2021.
\newblock \href {https://doi.org/10.48550/ARXIV.2104.07567} {Retrieval
  augmentation reduces hallucination in conversation}.

\bibitem[{Sun et~al.(2019)Sun, Yu, Chen, Yu, Choi, and Cardie}]{sun-2019-dream}
Kai Sun, Dian Yu, Jianshu Chen, Dong Yu, Yejin Choi, and Claire Cardie. 2019.
\newblock \href {https://doi.org/10.1162/tacl_a_00264} {{DREAM}: A challenge
  data set and models for dialogue-based reading comprehension}.
\newblock \emph{Transactions of the Association for Computational Linguistics},
  7:217--231.

\bibitem[{Tang et~al.(2022)Tang, Nair, Wang, Wang, Desai, Wade, Li,
  Celikyilmaz, Mehdad, and Radev}]{tang-etal-2022-confit}
Xiangru Tang, Arjun Nair, Borui Wang, Bingyao Wang, Jai Desai, Aaron Wade,
  Haoran Li, Asli Celikyilmaz, Yashar Mehdad, and Dragomir Radev. 2022.
\newblock \href {https://doi.org/10.18653/v1/2022.naacl-main.415} {{CONFIT}:
  Toward faithful dialogue summarization with linguistically-informed
  contrastive fine-tuning}.
\newblock In \emph{Proceedings of the 2022 Conference of the North American
  Chapter of the Association for Computational Linguistics: Human Language
  Technologies}, pages 5657--5668, Seattle, United States. Association for
  Computational Linguistics.

\bibitem[{Veličković et~al.(2018)Veličković, Cucurull, Casanova, Romero,
  Liò, and Bengio}]{velickovic2018graph}
Petar Veličković, Guillem Cucurull, Arantxa Casanova, Adriana Romero, Pietro
  Liò, and Yoshua Bengio. 2018.
\newblock \href {https://openreview.net/forum?id=rJXMpikCZ} {Graph attention
  networks}.
\newblock In \emph{International Conference on Learning Representations}.

\bibitem[{Wu et~al.(2021)Wu, Li, Xiao, Liu, Cao, Li, Wu, and Wang}]{wu2021bass}
Wenhao Wu, Wei Li, Xinyan Xiao, Jiachen Liu, Ziqiang Cao, Sujian Li, Hua Wu,
  and Haifeng Wang. 2021.
\newblock Bass: Boosting abstractive summarization with unified semantic graph.
\newblock \emph{arXiv preprint arXiv:2105.12041}.

\bibitem[{Yasunaga et~al.(2021)Yasunaga, Ren, Bosselut, Liang, and
  Leskovec}]{yasunaga-etal-2021-qa}
Michihiro Yasunaga, Hongyu Ren, Antoine Bosselut, Percy Liang, and Jure
  Leskovec. 2021.
\newblock \href {https://doi.org/10.18653/v1/2021.naacl-main.45} {{QA}-{GNN}:
  Reasoning with language models and knowledge graphs for question answering}.
\newblock In \emph{Proceedings of the 2021 Conference of the North American
  Chapter of the Association for Computational Linguistics: Human Language
  Technologies}, pages 535--546, Online. Association for Computational
  Linguistics.

\bibitem[{Zhang et~al.(2020)Zhang, Zhao, Saleh, and Liu}]{pmlr-v119-zhang20ae}
Jingqing Zhang, Yao Zhao, Mohammad Saleh, and Peter Liu. 2020.
\newblock \href {https://proceedings.mlr.press/v119/zhang20ae.html} {{PEGASUS}:
  Pre-training with extracted gap-sentences for abstractive summarization}.
\newblock In \emph{Proceedings of the 37th International Conference on Machine
  Learning}, volume 119 of \emph{Proceedings of Machine Learning Research},
  pages 11328--11339. PMLR.

\bibitem[{Zhang and Zhao(2021{\natexlab{a}})}]{zhang-2021-advances}
Zhuosheng Zhang and Hai Zhao. 2021{\natexlab{a}}.
\newblock \href {http://arxiv.org/abs/2103.03125} {Advances in multi-turn
  dialogue comprehension: A survey}.

\bibitem[{Zhang and Zhao(2021{\natexlab{b}})}]{zhang-zhao-2021-structural}
Zhuosheng Zhang and Hai Zhao. 2021{\natexlab{b}}.
\newblock \href {https://doi.org/10.18653/v1/2021.acl-long.399} {Structural
  pre-training for dialogue comprehension}.
\newblock In \emph{Proceedings of the 59th Annual Meeting of the Association
  for Computational Linguistics and the 11th International Joint Conference on
  Natural Language Processing (Volume 1: Long Papers)}, pages 5134--5145,
  Online. Association for Computational Linguistics.

\bibitem[{Zhong et~al.(2021)Zhong, Yin, Yu, Zaidi, Mutuma, Jha, Awadallah,
  Celikyilmaz, Liu, Qiu, and Radev}]{zhong-etal-2021-qmsum}
Ming Zhong, Da~Yin, Tao Yu, Ahmad Zaidi, Mutethia Mutuma, Rahul Jha,
  Ahmed~Hassan Awadallah, Asli Celikyilmaz, Yang Liu, Xipeng Qiu, and Dragomir
  Radev. 2021.
\newblock \href {https://doi.org/10.18653/v1/2021.naacl-main.472} {{QMS}um: A
  new benchmark for query-based multi-domain meeting summarization}.
\newblock In \emph{Proceedings of the 2021 Conference of the North American
  Chapter of the Association for Computational Linguistics: Human Language
  Technologies}, pages 5905--5921, Online. Association for Computational
  Linguistics.

\bibitem[{Zhu et~al.(2021)Zhu, Liu, Mei, and Zeng}]{zhu-etal-2021-mediasum}
Chenguang Zhu, Yang Liu, Jie Mei, and Michael Zeng. 2021.
\newblock \href {https://doi.org/10.18653/v1/2021.naacl-main.474}
  {{M}edia{S}um: A large-scale media interview dataset for dialogue
  summarization}.
\newblock In \emph{Proceedings of the 2021 Conference of the North American
  Chapter of the Association for Computational Linguistics: Human Language
  Technologies}, pages 5927--5934, Online. Association for Computational
  Linguistics.

\end{thebibliography}
\bibliographystyle{acl_natbib}

\end{document}